\documentclass[runningheads]{llncs}
\usepackage{graphicx}

\usepackage{tikz}
\usepackage{comment}
\usepackage{amsmath,amssymb} 
\usepackage{color}

\usepackage[accsupp]{axessibility}  

\usepackage{booktabs}
\usepackage{tabu}
\usepackage[strings]{underscore}

\begin{document}
\pagestyle{headings}
\mainmatter
\def\ECCVSubNumber{37}  

\title{Surgical Workflow Recognition: from Analysis of Challenges to Architectural Study} 

\titlerunning{Architectural Analysis for Surgical Workflow Recognition}
%
\author{Tobias Czempiel\inst{1} \and
Aidean Sharghi\inst{2}\and
Magdalini Paschali\inst{3}\and 
Nassir Navab\inst{1,4} \and
Omid Mohareri\inst{2}}
\authorrunning{T. Czempiel et al.}
%
\institute{Computer Aided Medical Procedures, Technical University of Munich, Germany \and
Intuitive Surgical Inc., Sunnyvale, USA \and 
Department of Psychiatry and Behavioral Sciences, Stanford University School of Medicine, Stanford, USA \and 
Computer Aided Medical Procedures, Johns Hopkins University, Baltimore, USA\\
}
\maketitle

\begin{abstract}
Algorithmic surgical workflow recognition is an ongoing research field and can be divided into laparoscopic (Internal) and operating room (External) analysis. So far, many different works for the internal analysis have been proposed with the combination of a frame-level and an additional temporal model to address the temporal ambiguities between different workflow phases. For the External recognition task, Clip-level methods are in the focus of researchers targeting the local ambiguities present in the operating room (OR) scene. 
In this work, we evaluate the performance of different combinations of common architectures for the task of surgical workflow recognition to provide a fair and comprehensive comparison of the methods for both settings, Internal and External. We show that the methods particularly designed for one setting can be transferred to the other mode and discuss the architecture effectiveness considering the main challenges for both Internal and External surgical workflow recognition.

\end{abstract}

\section{Introduction}

Automatic recognition of surgical workflow is essential for the Operating Room (OR) of the future, increasing the patient's safety through early detection of surgical workflow variations and improving the surgical results~\cite{Huaulme2020}. By supplying surgeons with the necessary information for each surgical step, a cognitive OR can reduce the stress induced by an overload of information~\cite{Katic2016} and build the foundation for more efficient surgical scheduling and reporting systems~\cite{berlet2022surgical}. 

Research in Surgical workflow analysis is separated into two fundamental direction (modes). The first mode is analyzing the internal surgical scene captured by laparoscopic or robotic cameras, which has been the main research focus in the past (Internal). Procedures such as laparoscopic cholecystectomy, colorectal and laparoscopic sleeve gastrectomy have been widely analyzed~\cite{Garrow2020}. 

The other mode is closely related to activity and action recognition, where cameras are placed inside the OR to capture human activities and external processes (External)~\cite{Srivastav2018}. These external cameras are rigidly installed to the ceiling or are attached to a portable cart. The goal of external OR workflow analysis is to capture the events happening inside the OR such as patient roll-in or docking of a surgical robot~\cite{Sharghi2020}.

Automatic surgical workflow recognition remains challenging due to the limited amount of publicly available annotated training data, visual similarities of different phases, and visual variability of frames among the same phase. The External OR scene is a complex environment with many surgical instruments and people working in a dense and cluttered environment. Compared to popular action recognition datasets, such as the Breakfast Action~\cite{Kuehne2014} or the GTEA~\cite{Fathi2011} dataset, the duration of a surgery is considerably longer. Therefore, the amount of information to be analyzed is substantially higher. Additionally, factors like variation of patient anatomy or surgeon style and personal preferences impose further challenges for the automatic analysis.

Our primary goal in this study is to identify the advantages and disadvantages of the fundamental building blocks used for surgical phase recognition in a structured and fair manner. We do not aim to establish a new state-of-the-art model or directly compare out-of-the-box methods, similar to endoscopic challenges. Our main objective and contribution is, considering the challenges of surgical phase recognition, to compare different network architecture blocks and main conceptual components that have been widely used for surgical phase recognition, analysing their advantages and disadvantages.

\subsection{Related Work} 
The analysis of the internal surgical workflow is an ongoing research topic~\cite{Padoy2012} that gained additional attention with the introduction of convolutional neural networks (CNNs) for computer vision tasks~\cite{Twinanda2017}. The basic building blocks of such approaches consisted of a frame-level method in combination with a temporal method to analyze the temporal context. Even though the frame-level methods got upgraded to more recent architectures~\cite{Bodenstedt2017,Jin2018} with more learning capabilities, the majority of research tried to improve the results with more capable temporal models such as Recurrent Neural Networks (RNNs)~\cite{Twinanda2017a}. Jin et al.~\cite{Jin2020} used Long short-term memory (LSTM) Networks~\cite{Bodenstedt2017} to temporally refine the surgical phase results with the prediction of the surgical tool and showed that both tasks profited from this combination. Czempiel et al.~\cite{10.1007/978-3-030-59716-0_33} proposed to replace the frequently used LSTMs with a multi-stage temporal convolution network (TCN)~\cite{Farha} analyzing the long temporal relationships more efficiently. Additionally, attention-based transformer architectures~\cite{Vaswani2017} have been proposed~\cite{Gao2021,czempiel2021opera} to refine the temporal context even further and increase model interpretability.  
\\

\begin{figure}[t] 
	\centering
	\includegraphics[width=.75\textwidth]{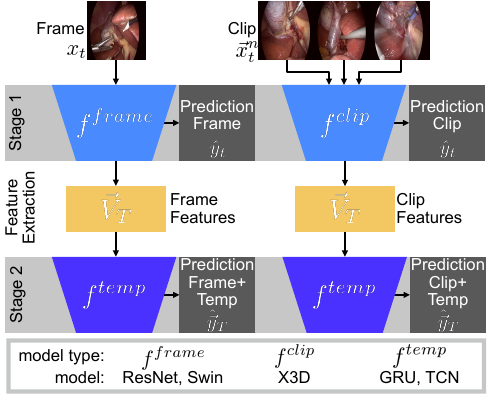}
	\caption{Stage 1 consists of the backbone models on both frame- and clip-level. Stage 2 describes the additional temporal training on extracted features}
	\label{fig2}
\end{figure}
\textbf{Fair Evaluation.}
One of the biggest challenges in this domain is the limited benchmarking between existing methods. New architectures are proposed, incrementally improving the results on the datasets. However, tracing improvements back to each fundamental architectural changes remains a challenging task. 

Researchers have to choose from a plethora of additional settings and methods that can influence the results such as data augmentation, choice of optimizers, learning rate schedules, or even dataset splits. The comparison gets even more challenging for 2-Stage methods that split the training into an image to feature and feature to prediction part which allows for long temporal modelling which otherwise would not be possible due to computational hardware limitations. In Figure~\ref{fig2} an overview of the different model combinations and configurations as Stage 1 and Stage 2 models is visualized. The novelty of some works is only limited to the Stage 2, however the quality of the extracted features from Stage 1 is essential for a method's overall prediction. Hence, the final results generated in the Stage 2 heavily rely on the Stage 1 features making it difficult to pinpoint the exact point of improvement. Finally, publicly available datasets usually provide limited training data, compromising model performance on unseen data. Thus, the conclusions we get can be misleading or only applicable to a particular dataset.

A very effective way to make a comparison fair for all participants is through the creation of challenges with an unpublished test set that is used to evaluate all submissions. The computer-assisted intervention community releases different challenges every year such as the Endovis\footnote{http://endovissub-workflow.grand-challenge.org} grand challenge. We strongly believe that challenges are a fundamental tool to identify the best solution for a surgical workflow task. However, even in public challenges the choice of learning strategies, learning rates, frameworks, optimizers vary among participants making the impact of each methodological advancement hard to identify. Nevertheless, architectural studies can provide additional insights to identify the methodological advancements in a structured way.

To this end, we conduct a fair and objective evaluation on multiple architectures for both Stages 1 and 2. We evaluate the various architectural components on two datasets for internal and external surgical workflow recognition.

\section{Analysis of Challenges: Local vs. Global}\label{challenges}
\begin{figure}[t] 
	\centering
	\includegraphics[width=.75\textwidth]{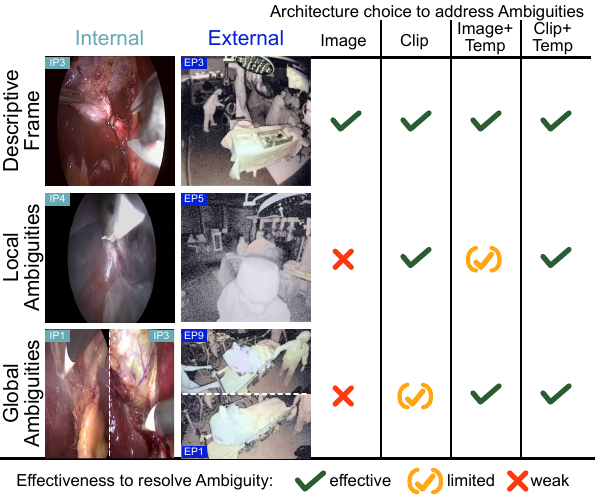}
	\caption{Different types of Surgical Workflow Datasets Internal (IE) and External (EP) along with three frames from each datasets that highlight Local and Global Ambiguities. On the right we highlight different architectural choices regarding the input type and addition of a temporal component (+Temp) that can be used to address the ambiguities}
	\label{fig1}
\end{figure}

 \textbf{Descriptive Frames.} In Fig.~\ref{fig1} 
 we visualized the four main architectural differences for video classification that 
 are discussed in this paper: end-to-end Image and Clip level methods and 2-Stage Image and Clip methods with an additional temporal model operating on extracted features. For descriptive frames such as the ones visualized in Figure~\ref{fig1} all of the aforementioned main architectural directions are likely to produce correct results. A single descriptive image is enough in this case to correctly identify the Phase as clipping\&cutting (IP3) mainly due to the presence of the clipping tool that only appears in that particular phase.

\noindent\textbf{Local Ambiguities.} In the second row of Figure~\ref{fig1} two examples of frames with Local Ambiguities are shown. In the external scene several occlusions of the camera appear, caused by bulky OR equipment or medical staff operating in a crowded environment. In the internal scene, the view can be impeded by smoke generated during tissue coagulation or pollution of the endoscopic lens through body fluids. Surgical frames do not receive a phase label purely based on their visual properties but also based on their semantic meaning within the OR. 
Surgical workflow analysis methods should be able to categorize ambiguous frames, such as ones captured with a polluted lens, based on this global semantic context.
However, purely image-level methods lack the temporal context and the semantic information of previous time points to resolve ambiguities reliably. 
Clip-level methods are able to understand the context of a frame neighbourhood, which is mostly sufficient as these Local Ambiguities in the internal and external OR scene often only persist for a few seconds before the person in the scene moves to a different location or the surgeon cleans the lens to continue the surgery safely.

\noindent\textbf{Global Ambiguities.}
There are many activities in the external and internal analysis with high visual similarities belonging to different phases (high inter-class similarities). For instance, the phases of patient roll-in and roll-out in the external dataset contain frames that look almost identical (Figure~\ref{fig1}). To correctly differentiate these two phases only based on single images could be challenging even for an experience surgeon. Considering a limited temporal context could also be insufficient for a correct classification, since the activity can be considerably longer. However, the rich temporal context modeled in a 2-Stage approach, can alleviate this confusion more effectively as it is clear that the roll-out phase appears always after the roll-in phase.

\section{Methodology}
We conduct a fair and objective evaluation on multiple architectures under the same settings. We evaluate our results on two different datasets for internal and external surgical workflow recognition.
Each surgical workflow video consists of different amount of frames $T$. For each frame $x_t \in \{x_1, x_2, \ldots x_T\}$ the target is to create an accurate prediction corresponding to the label ${y_t}$ of the frame-level model $f^{frame}(x_{t}) = \hat{y_t}$. For the clip-level model the input is not a single frame but a sequence of frames $\vec{x}^{n}_{t} = \left(x_{t-n}, x_{t-n+1}, \ldots , x_{t} \right)$ were $n$ is the size of the input clip to generate one prediction $f^{clip}\left(\vec{x}^{n}_{t}\right) = \hat{y_t}$.
In the feature extraction process we extract for each time point $t$ of a video a feature vector $\vec{v}_t$. The feature vectors from an entire video $\vec{V}_T = \{ \vec{v}_1, \vec{v}_2, \ldots, \vec{v}_T \}$ are then used as input for the temporal models of the second Stage that predict a corresponding output probability for all feature vectors  $f^{temp}(\vec{V}_T) = \hat{\vec{y}}_T$, where $\hat{\vec{y}}_T = ( \hat{y_1},  \hat{y_2}, \ldots, \hat{y_T}  )$.

\subsection{Visual Backbone - Stage 1}\label{visualbackbone}
\textbf{Frame-level.} Frame-level methods operate on a single image using the visual contend to create a classification (Figure~\ref{fig2} - Prediction Frame). For the task of surgical workflow recognition a purely frame-level model has problems in resolving Local Ambiguities as discussed in section~\ref{challenges}. 
The one exception to this is when a particular local ambiguity only appears in a single phase of the surgery which would make the local ambiguity a descriptive frame for both algorithm and clinician.
A surgeon would still be able to resolve local ambiguities by considering the frames proceeding the ambiguous one. 
\\\textbf{Clip-level.} We can imitate the strategy to also consider proceeding frames using Clip-level models that take as input multiple frames to refine the frame-level prediction. Clip-level models usually have a limited temporal knowledge but have the ability to detect motion patterns and directions which is critical when predicting activities. In contrast to frame-level models, clip-level models are more resilient to local ambiguities by design. 
However, both frame and clip level models have trouble handling Global Ambiguities. Thus, using a Temporal Model in an additional training Stage can be used to address this challenge.
\\\textbf{Models.} For the frame-level methods we build up on the many previous works in this domain by using the CNN based ResNet-50~\cite{He2016} model. Additionally, we utilize a novel visual transformer network Swin~\cite{Liu2021} into our study, that does not perform any convolutions but purely relies on self-attention, with state-of-the-art performance on the large scale Imagenet image classifcation task~\cite{deng2009imagenet}. 
Both models have a similar amount of parameters and memory requirements. For the Clip-level methods we choose the expanding architectures for efficient video recognition (X3D)~\cite{Feichtenhofer2020} based on 3D CNNs and residual layers, which outperform many methods with a substantially higher number of learnable parameter on the challenging kinetics human action dataset~\cite{Smaira2020}.

\noindent\textbf{Feature Extraction.} 
Due to hardware memory limitations, training the Visual Backbone end-to-end with a temporal model is often impossible. Hence, training the temporal models is done in a second Stage using the feature embedding $v$ extracted from the Stage 1 models.

\subsection{Temporal Models - Stage 2}\label{temporalmodels}

To address the Global Ambiguities discussed in section~\ref{challenges} an additional temporal model can be used with the capabilities to analyse the entire surgical procedure at once. 
\\\textbf{Models}. For the temporal models of the Stage 2 we first utilize commonly used architectures, such as RNNs and specifically Gated Recurrent Units (GRU)~\cite{Bahdanau2015} which have been shown to achieve comparable results to LSTMs with a simplified architecture \cite{Bodenstedt2019}. 
RNNs follow an autoregressive pattern for the inference where each prediction builds up on the previous one.
Temporal convolutions~\cite{Farha} (TCN), on the other hand, follow a non-autoregressive structure taking all the past time points as input for the current prediction. Temporal transformer architectures follow a similar non-autoregressive pattern but tend to overfit more easily due to their increase amount of learable parameters. TCNs are lightweight and fast to train which is why we selected them as the architectural choice for the non-autoregressive group. In Figure~\ref{fig2} the different combinations of Stage 1 frame- and clip-based models with a Stage 2 temporal model is visualized.

\section{Experimental Setup}
In the following we conducted an ablation of the aforementioned methods while keeping the hyperparameters of the architecture comparable to conduct a fair and comprehensive study on two dataset for surgical workflow recognition. The phase names of both datasets are summarized in Table~\ref{tab:Phase Definitions}. The Cholec80 dataset contains surgery specific phases while the External dataset phases are less specific to an intervention.

\subsection{Datasets}
\begin{table}[t]
    \centering
    \caption{Phase Definitions for the datasets used in this study Cholec80 as the Internal dataset with Internal Phases (IP) and External DS as External dataset with External Phases (EP).}
    \label{tab:Phase Definitions}
    \begin{tabu} to \textwidth { X[c] X[3c] X[c] X[3c] }
    \multicolumn{2}{c}{\textbf{Cholec80 Phase Names}} &    \multicolumn{2}{c}{\textbf{External DS Phase Names}}  \\
    \cmidrule(lr){1-2} \cmidrule(lr){3-4} 
    IP1 & preparation & EP1 &  sterile preparation\\
    IP2 & calot triangle dissection & EP2  &  patient roll-in\\
    IP3 & clipping\&cutting & EP3  &  patient preparation \\
    IP4 & gallbladder dissection & EP4  &  robot roll-up  \\
    IP5 & gallbladder packaging & EP5  &  robot docking  \\
    IP6 & cleaning \& coagulation & EP6  &  surgery \\
    IP7 & gallbladder retraction & EP7  &  robot undocking \\
     &  & EP8  &  robot rollback \\
     &  & EP9  &  patient close \\
     &  & EP10 &  patient roll-out \\
    \end{tabu}
\end{table}

\textbf{Cholec80.} The internal workflow recognition capabilities are evaluated with the publicly available Cholec80~\cite{Twinanda2017} dataset that includes 80 cholecystectomy surgeries were every RGB-frame belongs to one out of seven surgical phases (Table~\ref{tab:Phase Definitions}). We split the dataset into half training and half testing as described in \cite{Twinanda2017} For a fair comparison between datasets we did not use the additional surgical tool labels in Cholec80. \\
\textbf{External DS.} The external dataset (External DS) includes 400 videos from 103 robotic surgical cases. The dataset includes 10 different phases describing the workflow in the OR (Table~\ref{tab:Phase Definitions}). Every surgery is recorded from four different angles. We treat each angle as a separate input for our model and strictly separate the recordings on a procedural level. Recordings of the same surgery from different views were always in the same train or test split. 
In this work, we do not consider the combination of multiple views of one surgery as this is out of scope for our work. We used time of flight cameras (ToF) not only for improved privacy compared to RGB but ToF also provides geometrical rich information about the scene. External DS contains 10 general, surgery type agnostic, classes for robotic surgery in a multi-label setting. The dataset contains 28 different surgery types all recorded using an daVinci Xi surgical system. 
We used 80\% of all videos for training and the remaining 20\% of all videos for testing. To ensure generalizability of our approach to various minimally invasive surgical procedures, all training videos originated from 16 surgery types and all testing videos from the remaining 12 surgery types with no overlap.

\noindent \textbf{Model Training.}
For our training setup we keep the training configuration options constant between the different methods to establish a fair evaluation. This way, we try to develop a comparison focusing on the main architectural design choices without unwanted metric advantages by using, e.g., a more sophisticated optimizer or training scheme. For the learning on the multi-class Cholec80 dataset we are using the Cross-Entropy and the Binary Cross-Entropy loss for the multi-label External DS. For all our experiments we used the Adam optimizer with a learning rate of 1e-4 for the Stage 1 models and 1e-3 for the Stage 2 models with a step learing rate scheduler (beta: 0.1, interval: 10 epochs). For the training of the Stage 1 models we used the RandAugment \cite{Cubuk2020} which combines a multitude of different augmentation techniques in an optimized manner. For both datasets, we resize all the frames from each video to 224x224 pixels which allowed us to use pretrained weights on ImageNet \cite{deng2009imagenet} (ResNet, Swin) and Kinetics \cite{Kay2017} (X3D) to accelerate the convergence. We further follow the related work and subsample both datasets to 1 frame per second. All of our experiments are performed using python and the deep learning training library pytorch. 
\subsection{Architecture Settings} 
For all our models we adapted the size of the output fully-connected layer to match the number of surgical phases for each dataset (Output layer - Internal dataset: 7, Output layer - External dataset: 10).
\\\textbf{ResNet.} We choose the ResNet-50 model with pretrained ImageNet weights.
\\\textbf{Swin.} For the visual transformer architecture we choose the Swin-T version with initialized weights from Imagenet.
\\\textbf{X3D.} For the X3D architecture we choose X3D-M a efficient compromise between model size and performance. We set the size of the input clip $n$ to 16 frames resulting in a temporal receptive window of 16 seconds. The weights are initialized from Kinetics. \\\textbf{GRU} For GRU we used a hidden dimensionality matching the dimensionality of the features extrated from the Stage 1 models. Additionally  we selected 2 GRU layers.
\\\textbf{TCN.} For the TCN model we used 15 layers over 2 Stages and ensured to set the models in causal inference mode for online phase recognition results without temporal leakage. \\
\subsection{Evaluation Metrics and Baselines.}
To comprehensively measure the results we report the the harmonic mean (F1) of Precision and Recall~\cite{Padoy2012}. Precision and Recall are calculated by averaging the results for each class over all samples followed by an average over all class averages. In that way the metrics are reported on a video level to make sure that a long and short intervention contribute equally to the results. For the multiclass cholec80 dataset we also used the mean accuracy following Endonet \cite{Twinanda2017}, averaged over all videos of the test set, since it is a commonly used metric for surgical workflow recognition. For the multi-label external surgical workflow dataset we chose the mean average precision (mAP \cite{idrees2017thumos}) as an objective metric similar to other works in the field of activity and action recognition \cite{Sharghi2020}. All of the models are run with 3 different random seeds and the mean and standard variation across the runs are reported. 

\section{Results and Discussion} \label{results}
In Table~\ref{ComparisonTable} the models with temporal components in Stage 2 (GRU, TCN) have been trained with extracted features from the respective visual backbones of Stage 1 (ResNet, Swin, X3D). On both Cholec80 and External DS the image-level Swin backbone improves the results of the ResNet architecture on all metrics. The clip-level modeling with local temporal context in X3D further improves the results which is especially noticeable in the F1 score for both Cholec80 (+5.69\%) and the External DS (+8.62\%). 

Furthermore, with the addition of a temporal model we can refine the results, leading in an increase in both Accuracy and F1 on Cholec80 for both the TCN and GRU. On Cholec80, the TCN improves the results more than GRU on most settings (ResNet, X3D) but the combination of Swin and GRU achives the highest accuracy. For the External DS although the F1 results are generally more stable with the addition of TCN, the best result is achieved with a combination of X3D with GRU (80.35 mAP, 76.70 F1).

\begin{table}[t]
    \centering
    \caption{Comparative study of architectural components for surgical phase recognition using different model backbones for Stages 1 and 2 on our selected surgical workflow datasets.}
    \label{ComparisonTable}
    \begin{tabu} to \textwidth { X[c] X[c] X[1.8c] X[1.8c] X[1.8c] X[1.8c]}
     \multicolumn{2}{c}{\textbf{Architecture}} & \multicolumn{2}{c}{\textbf{Cholec80}} & \multicolumn{2}{c}{\textbf{External DS}} \\ 
     Stage 1 & Stage 2 & \centering{Acc} & \centering{F1} & \centering{mAP} & \centering{F1}  \\
    \cmidrule(lr){1-2} \cmidrule(lr){3-4} \cmidrule(lr){5-6}
    ResNet & & 81.23$\pm$2.5 & 71.27$\pm$3.0 & 67.50$\pm$1.9 & 58.66$\pm$0.7 \\
    Swin & & 82.16$\pm$1.7 & 71.86$\pm$2.3 & 68.89$\pm$1.8 & 59.89$\pm$0.4 \\
    X3D & & 82.62$\pm$2.6 & 77.55$\pm$1.8 & 69.33$\pm$1.4 & 68.42$\pm$0.9 \\
    \midrule
    ResNet & GRU & 85.62$\pm$2.4 & 76.68$\pm$1.7 & 70.8$\pm$1.9 & 65.12$\pm$1.5 \\
    Swin & GRU & \textbf{87.73}$\pm$\textbf{2.2} & 80.65$\pm$1.2 & 67.81$\pm$2.0 & 63.65$\pm$1.3 \\
    X3D & GRU & 85.53$\pm$1.7 & 78.99$\pm$1.4 & \textbf{80.35}$\pm$\textbf{1.5} & \textbf{76.70}$\pm$\textbf{1.1} \\
    \midrule
    ResNet & TCN & 87.37$\pm$1.4 & 82.48$\pm$1.8 & 67.34$\pm$1.9 & 65.35$\pm$0.9 \\
    Swin & TCN & 87.55$\pm$0.7 & \textbf{81.70}$\pm$\textbf{1.6} & 73.02$\pm$3.1 & 71.08$\pm$2.0 \\
    X3D & TCN & 85.81$\pm$1.4 & 80.41$\pm$1.3 & 77.15$\pm$1.5 & 74.96$\pm$1.6 \\
    \end{tabu}
    \label{tab:baseline_testing}
\end{table}


%




In the discussion we want to bring together the implications of the Analysis of Challenges (Section~\ref{challenges}) and the practical results of the architectural study. In Figure~\ref{fig2} we see that the addition of the temporal models should indeed improve the results as it is more capable to establish a longer temporal context and resolving the global ambiguities.
In fact, the results (section~\ref{results}) showcase that the addition of a temporal model such as GRU or TCN improves the metrics across the board. 

Additionally, as we reasoned in Section~\ref{challenges}, Clip-level models would be able to resolve Local Ambiguities more effectively in comparison to Image-level models. Indeed, our results highlight that the clip-level model achieves higher Acc (82.62) compared to both image-level models and outperforms them regarding their F1-score on both datasets (+6\% and +8\%). 

Interestingly, all metrics on Cholec80 highlight that the addition of a temporal model is beneficial, as expected. However, for External DS SwinGRU and ResTCN achieve marginally lower mAPs (67.81 and 67.34 respectively) compared to Swin (68.89) and ResNet (67.50). However, the F1-score for the External DS is still consistently improved by the addition of a Stage 2 model, which shows the need for evaluating multiple metrics to determine which model would be better suited for computer-aided surgical workflow systems.

Moreover, our results show that the clip-level backbone could not reach the performance of the frame-level backbones when combined with a temporal model on Cholec80. 
This could be attributed to the limited amount of training data in comparison to External DS and to the increased amount of learnable parameters, that could cause overfitting on the clip-level backbone. We believe, that since clip-level models are capable of overcoming Local Ambiguities and have shown to outperform image-level architectures on the External DS, they will be more widely used and preferred once the internal surgical workflow recognition datasets, as soon as they expand further.

\section{Conclusion}
In this work we presented a fair analysis of models for the task of surgical workflow recognition on two datasets. We conducted an analysis of the challenges related to surgical workflow recognition and provided an intuition on how architectures can be chosen specifically to address them. 

Our results show that utilizing a Temporal Model is a critical component of a model for surgical workflow and can help overcome Global Ambiguities. TCNs are particularly suitable as temporal models, combining a low number of trainable parameters and a large receptive field. Furthermore, we showed that clip-level models can alleviate Local Ambiguities on the External DS and have the potential of benefiting Internal Datasets as they expand in training data size. 
Future work includes utilizing recently introduced transformer architectures~\cite{czempiel2021opera,Gao2021} as temporal backbones, since they have the potential to increase the model performance. However, using larger datasets is critical when training such complex architectures or analyzing procedures including more surgical phases with longer duration and higher surgical complexity.

Furthermore, as has been recently shown~\cite{reinke2022metrics} the choice of the evaluation metric is crucial and should carefully consider for each task. This is in line with the results of our study, where models' mAP, Acc and F1-score were critical for the fair evaluation and comparison of each architecture. Finally, when comparing model architectures for surgical workflow analysis, the model robustness~\cite{paschali2018generalizability} and tolerance to outliers~\cite{berger2021confidence} could additionally be taken into account.

\medskip
\noindent \textbf{Acknowledgements.}
The majority of this work was carried out during an internship by Tobias Czempiel at Intuitive Surgical Inc.

\clearpage
%
%

\bibliographystyle{splncs04}
\bibliography{bibfile}
\end{document}